\documentclass[review]{elsarticle}

\usepackage{lineno,hyperref}
\usepackage{amsmath}
\DeclareMathOperator*{\argmax}{arg\,max}
\usepackage{amssymb}
\usepackage{caption}
\usepackage{subcaption}
\usepackage{booktabs}
\usepackage{xcolor}

\newcommand{\bcheck}{\color{blue}{\bf \checkmark}}
\usepackage{lineno}
\journal{Ocean Engineering}

\begin{document}

\begin{frontmatter}

\title{Prediction-Free, Real-Time Flexible Control of Tidal Lagoons through Proximal Policy Optimisation: A Case Study for the Swansea Lagoon}
% \tnotetext[mytitlenote]{Fully documented templates are available in the elsarticle package on \href{http://www.ctan.org/tex-archive/macros/latex/contrib/elsarticle}{CTAN}.}

\author{Túlio Marcondes Moreira\corref{mycorrespondingauthor}$^a$}
% \fntext[myfootnote]{Since 1880.} % Extra footnotes can be added with \fnref{myfootnote}
% \ead{tuliommoreira.tm@gmail.com}

\author{Jackson Geraldo de Faria Jr$^a$}

\author{Pedro O.S. Vaz de Melo$^a$}

\author{Luiz Chaimowicz$^a$}

\author{Gilberto Medeiros-Ribeiro$^a$}

\address{$^a$Computer Science Department (DCC), Universidade Federal de Minas Gerais, Belo Horizonte, Minas Gerais, 31270-901, Brazil}

% \author[mymainaddress,mysecondaryaddress]{Elsevier Inc}

% \author[mysecondaryaddress]{Global Customer Service}
\cortext[mycorrespondingauthor]{Corresponding author e-mail address: tuliommoreira.tm@gmail.com}

% \address[mymainaddress]{1600 John F Kennedy Boulevard, Philadelphia}
% \address[mysecondaryaddress]{360 Park Avenue South, New York}

\begin{abstract}
Tidal Range Structures (TRS) have been considered for large-scale electricity generation for their potential ability to produce reasonably predictable energy without the emission of greenhouse gases. Once the main forcing components for driving the tides have deterministic dynamics, the available energy in a given TRS has been estimated, through analytical and numerical optimisation routines, as a mostly predictable event. This constraint imposes state-of-art flexible operation methods to rely on tidal predictions to infer best operational strategies for TRS, with the additional cost of requiring to run optimisation routines for every new tide. In this paper, a Deep Reinforcement Learning approach (Proximal Policy Optimisation through Unity ML-Agents) is introduced to perform automatic operation of TRS. For validation, the performance of the proposed method is compared with six different operation optimisation approaches devised from the literature, utilising the Swansea Bay Tidal Lagoon as a case study. We show that our approach is successful in maximising energy generation through an optimised operational policy of turbines and sluices, yielding competitive results with state-of-art optimisation strategies, with the clear advantages of requiring training once and performing real-time automatic control of TRS with measured ocean data only.
\end{abstract}


\end{frontmatter} % Reduce Abstract to remove error output

% \linenumbers

\section{Introduction}

In recent years, concerns about climate  change combined with political and social pressures have pushed the world to increase the installed capacity of renewable energy sources (wind,  solar,  bioenergy and hydro), allowing renewable energy to account for 28\% of global energy generation in 2020 \cite{globalEn}. While significant progress has been made in expanding solar and wind resources, tidal energy remains practically untapped. As of today, only two successful large Tidal Range Structure (TRS) projects have been built, namely, La Rance (France) and Lake Sihwa (South Korea), with 240 $MW$ and 254 $MW$ of installed capacity, respectively \cite{neill2018tidal}. 

A review by UK's ex-minister of energy \cite{hendry2016role} has drawn attention to TRS as a  competitive  choice  among renewables. In his report, the construction of ``small-scale'' TRS, such as the Swansea Bay Tidal Lagoon (our case study), is suggested as a pathfinder project before moving to larger-scale TRS \cite{waters2016world, petley2016swansea, burrows2009tidal}. The report also emphasises that TRS have proposed lifetimes of operation of 120 years -- far surpassing any other renewable energy type, allowing for very low electricity cost for years. As an example, La Rance, which is in operation for 55 years, took 20 years to amortise the initial investment, generating energy at competitive cost of nuclear or offshore wind sources \cite{LaRance20, hendry2016role}. 

Among the challenges faced in TRS deployment is the optimisation of energy generation through the operation of hydraulic structures (turbines and sluices), which increases the utilisation factor (ratio of actual energy generated to installed capacity). Although the current literature has advanced in increasing theoretical power generation capabilities of TRS \cite{neill2018tidal, waters2016tidal, ghaedi2021generated}, there is room for improvement, considering that state-of-art (flexible operation) optimisation methods (i) can be computationally time expensive and (ii) do not perform real-time control, relying on accurate tidal prediction techniques. In view of this, we propose the usage of Deep Reinforcement Learning (DRL) methods, more specifically Proximal Policy Optimisation (PPO) through the Unity ML-Agents package \cite{juliani2018unity}, which enables state-of-art energy generation of TRS (on par with best optimisation routines) through the real-time control of turbines and sluices. DRL was chosen among machine learning techniques due to the nature of our problem, which involves sequential decision-making of a reactive environment (lagoon water levels vary depending on the operation of hydraulic structures) with the goal of maximising expected return (energy), and also because a target optimal operation of the tidal lagoon is not known ``a priori'' -- a requirement for supervised learning techniques. After training, our method shows consistent performance, regardless of test data used, not requiring future tidal predictions or re-training the DRL agent. To date, this is the first flexible operation optimisation approach in the literature that can maximise TRS energy generation without such constraints. A 0D model of the Swansea Bay Tidal Lagoon is utilised to compare our DRL method with six optimisation baselines devised from the literature \cite{angeloudis2018optimising, xue2019optimising}.

After related work (Section \ref{relatedw}), this paper is divided in four parts. In the first part, we explain how tidal barrages extract energy from the tides through the lenses of classical and variant operation approaches. In the second part, we cover the theory behind DRL, more specifically on the PPO algorithm that is used in this study. In the third part, we cover our agent-environment setup, modelled with the Unity ML-Agents package (Table S1, Supplementary Material). In the final part, an experimental study contrasting our results with six baselines is presented and discussed.

\section{Related Work} \label{relatedw}

State-of-art optimisation methods for TRS estimate the available energy in such systems by operating the tidal lagoon hydraulic structures through flexible operational strategies (adaptive, according to tidal amplitudes and lagoon water levels for sequential tidal cycles \cite{xue2019optimising,angeloudis2018optimising}). With the assumption of well predictable tides, flexible operation of turbines and sluices can be inferred by ``looking--ahead'' through harmonic or numerical tidal prediction methods \cite{egbert2017tidal} and applying the acquired operation to the real, measured ocean -- a procedure that needs to be repeated for every new tide. In fact, and to the best of our knowledge, the requirement of accurate future tidal predictions (concurrent with measured data and up to a multiple of half-tidal cycles into the future) has been the basis for all optimisation routines developed for enabling flexible operation \cite{ahmadian2017optimisation, xue2019optimising, angeloudis2018optimising, xue2020genetic, harcourt2019utilising,neill2018tidal, xue2021design}. This constraint can be a problem when future tidal predictions are unavailable, unreliable, have some associated validation cost \cite{medina2021satellite} or are regulated by private companies or government agencies. 

% Additionally, even if tidal predictions are available and reliable, annual Operation and Maintenance costs (O\&M), equating to $\approx 1.75\%$ of the total TRS construction cost \cite{hammond2017technology}, are expected when state-of-art optimisation methods are employed. 

State-of-art optimisation routines in the literature utilise either grid search (a brute-force approach), gradient-based, global optimisation \cite{xue2019optimising, angeloudis2018optimising, harcourt2019utilising} and, more recently, genetic algorithm methods \cite{xue2021design} to optimise the operation of TRS. As a basis of comparison with our DRL agent, two non-flexible and four flexible state-of-art baselines devised from \cite{angeloudis2018optimising, xue2019optimising} are modelled  utilising grid search and global optimisation methods.

% For instance, research show that brute-force optimisation of starting ($H_{start}$) and finishing ($H_{min}$) turbine operational heads (water level difference between ocean and lagoon) for every consecutive tide in a two-way scheme, could lead to significant increases in power output. This method was applied to the Swansea Bay Tidal Lagoon \cite{ahmadian2017optimisation}, leading to a 15\% increase in the electricity generation over a year, in contrast to usual optimisation strategies with constant turbine operational head values. However, this result assumed that all tide signals in a year were known ``a priori'', so that $H_{start}$ and $H_{min}$ could be optimised for each sequential tide. 
% In this work, six state 

\section{Tidal Power Overview} \label{TidalPower}

\begin{figure}
	\centering
	\includegraphics[width=.8\linewidth]{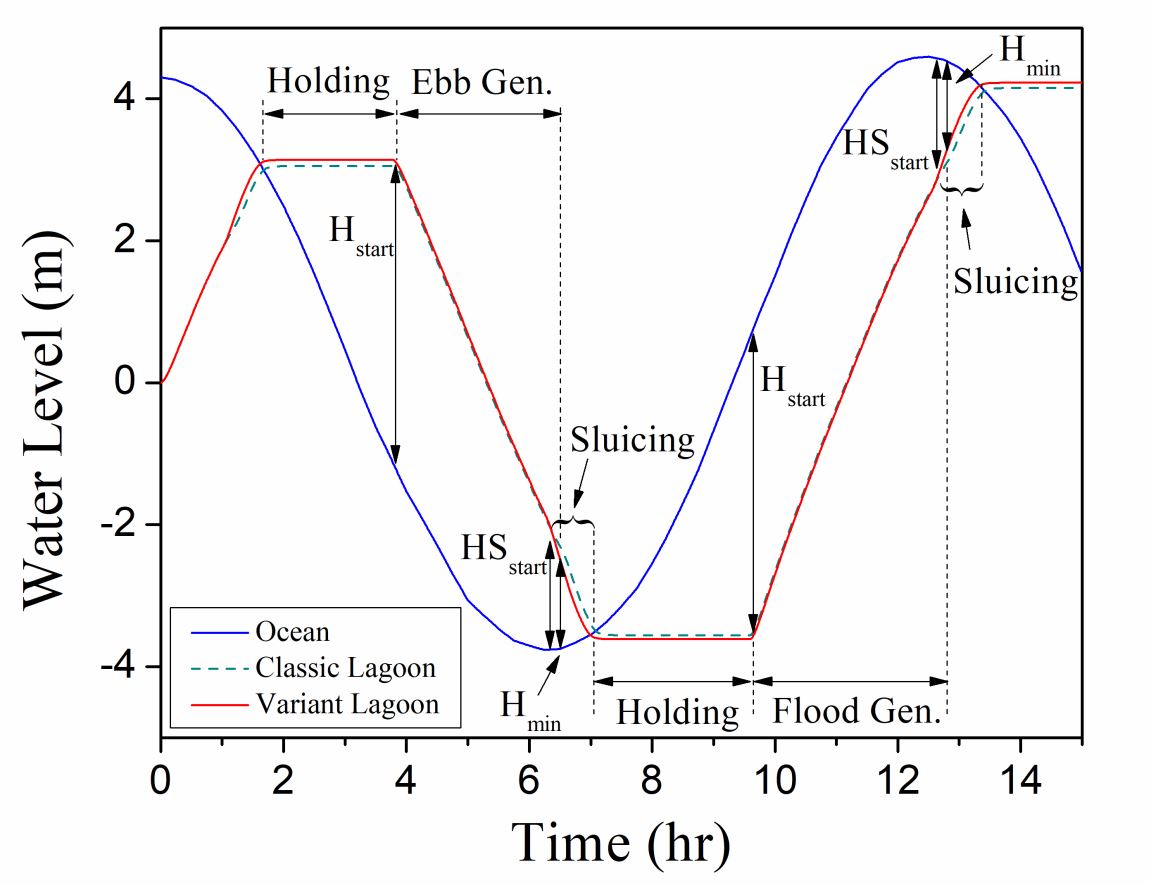}
	\caption{Classic and variant ``two-way scheme'' operation. Ocean level is represented by the blue line, while the lagoon level is shown in either green dashed lines or red, for classic or variant lagoon operations, respectively.}\label{Operation}
\end{figure}

TRS extract power by artificially inducing a water head difference between the ocean and an impounded area. By allowing water to flow through the hydraulic structures into an artificial impoundment, the incoming tide (flood tide) is confined within the lagoon at high level (holding stage). Then, during the receding tide (ebb tide), power generation begins when a high operational head ($H_{start}$) is established between the basin and ocean \cite{prandle1984simple}. Power generation then stops when a minimum operational head ($H_{min}$) is achieved. A sluicing sequence immediately follows, where idling turbines and sluices allow water to flow in order to increase lagoon tidal range for the next operation. Following the same procedure, generating energy is also possible during the flood tide, although with reduced efficiency due to turbines usually being ebb-oriented \cite{angeloudis2018optimising}. From the literature, operational strategies that allow for power generation to occur during flood and ebb tides are called ``two-way scheme'' operation \cite{prandle1984simple}.  Operational modes for controlling TRS in a ``two-way scheme'' are shown in Fig. \ref{Operation} and detailed in Table \ref{Tab1}.

\begin{table}
\centering
  \caption{TRS control stages.}\label{Tab1}
  \begin{tabular}{ccl}
    \toprule
    Operational Mode & Description\\
    \midrule
    Ebb Gen:  & Power generation during receding tide \\
    Flood Gen:  & Power generation during incoming tide \\
    Sluicing:  & Operate sluice gates and/or idle turbines \\
    Holding:  & Stop operation of all hydraulic structures \\
  \bottomrule
\end{tabular}
\end{table}

\begin{table}
\centering
  \caption{Classic hydraulic structures operation.}\label{Tab2}
  \begin{tabular}{cccl}
    \toprule
    Operational Mode & Turbines & Sluices & Power Gen. \\
    \midrule
    Ebb Gen:    & On & Off & Yes (if $H_{start} > H_{mt}$) \\
    Flood Gen:  & On & Off & Yes (if $H_{start} > H_{mt}$)\\
    Sluicing:   & On & On & No\\
    Holding:    & Off & Off & No\\
  \bottomrule
\end{tabular}
\end{table}

TRS turbines can be operated to either generate energy or to increase flow rates through the barrage during sluicing stage (idle operation of turbines). Also, a minimum head $H_{mt}$, usually in the $[\text{1 }m - \text{2 }m]$ range \cite{aggidis2013operational}, is required for the turbine to generate energy ($H_{mt} = \text{1 }m$ in this study). Considering that the holding stage begins automatically when the difference between ocean and lagoon is negligible, and that power generation is not possible with head differences below $H_{mt}$, the classic operation of tidal lagoons \cite{prandle1984simple} is reduced to two variables: $H_{start}$ and $H_{min}$. As seen in Fig. \ref{Operation}, pairs $H_{start}$ and $H_{min}$ occur every half-tide period, when ocean oscillates between its valleys and peaks. 

A slight modification of the discussed classical operation allows for opening the sluice gates at the end of ``flood'' and ``ebb'' generation stages \cite{Baker, angeloudis2018optimising} independently of $H_{min}$, with the possibility of increasing power generation (increased lagoon tidal range when starting the next ``ebb'' or ``flood'' stages). This variant operation requires 3 control variables every half-tide: $H_{start}$, $H_{min}$ and $HS_{start}$ (sluice gate starting head). The water level variations within the lagoon, following classic and variant operations of hydraulic structures, can be seen in Fig. \ref{Operation}. Tables \ref{Tab2} and \ref{VariantOpTable} show all possible combined operations of turbines and sluices, with resulting power generation, for each control stage in classic and variant operations, respectively. Classic and variant approaches to operate TRS are used in the optimisation routines of our baselines in Section \ref{baselines}.

\begin{table}
\centering
\setlength{\tabcolsep}{4pt}
  \caption{Variant hydraulic structures operation.}\label{VariantOpTable}
  \begin{tabular}{cccl}
    \toprule
    Operational Mode & Turbines & Sluices & Power Gen. \\
    \midrule
    Ebb Gen:    & On & Off/On & Yes (if $H_{start} > H_{mt}$) \\
    Flood Gen:  & On & Off/On & Yes (if $H_{start} > H_{mt}$)\\
    Sluicing:   & On & Off/On & No\\
    Holding:    & Off & Off & No\\
  \bottomrule
\end{tabular}
\end{table}

\subsection{Tidal Lagoon Simulation - 0D Model} \label{0DModel}

In order to estimate the available energy of two-way operational strategies, analytical or numerical models (0D to 3D) can be considered. When the goal is the optimisation of TRS operation for maximising energy generation, 0D models are usually chosen, given their computational efficiency, and the fact that for ``small-scale'' projects, such as the Swansea Bay Tidal Lagoon, 0D models present good agreement with more complex finite-element 2D models \cite{angeloudis2017comparison, angeloudis2018optimising, xue2019optimising, neill2018tidal, andrea2019implementation}. 0D models are derived from conservation of mass:
\begin{equation} 
  \frac{d L}{d t} = \frac{Q_T}{Al(L)},
  \label{0D}
\end{equation}
where $L$ is the water level (in meters) inside the lagoon, $Q_T$ is the total directional water flow rate ($m^3/s$) from both sluices and turbines and $Al(L)$ is the variable lagoon area ($m^2$). From Eq.~\ref{0D}, the lagoon water level at the following time-step ($L_{t+1}$) can be calculated by a backward finite difference method:
\begin{equation} 
  L_{t+1} = L_t + \frac{Q_T}{Al(L)} \Delta t,
  \label{BD}
\end{equation}
where $L_{t}$ is the water level at time-step $t$ and $\Delta t$ the discretized time ($60~s$ in this work).

\subsection{Turbine and Sluice Parametrization} \label{Equations}

\begin{figure}
	\centering
	\includegraphics[width=.8\linewidth]{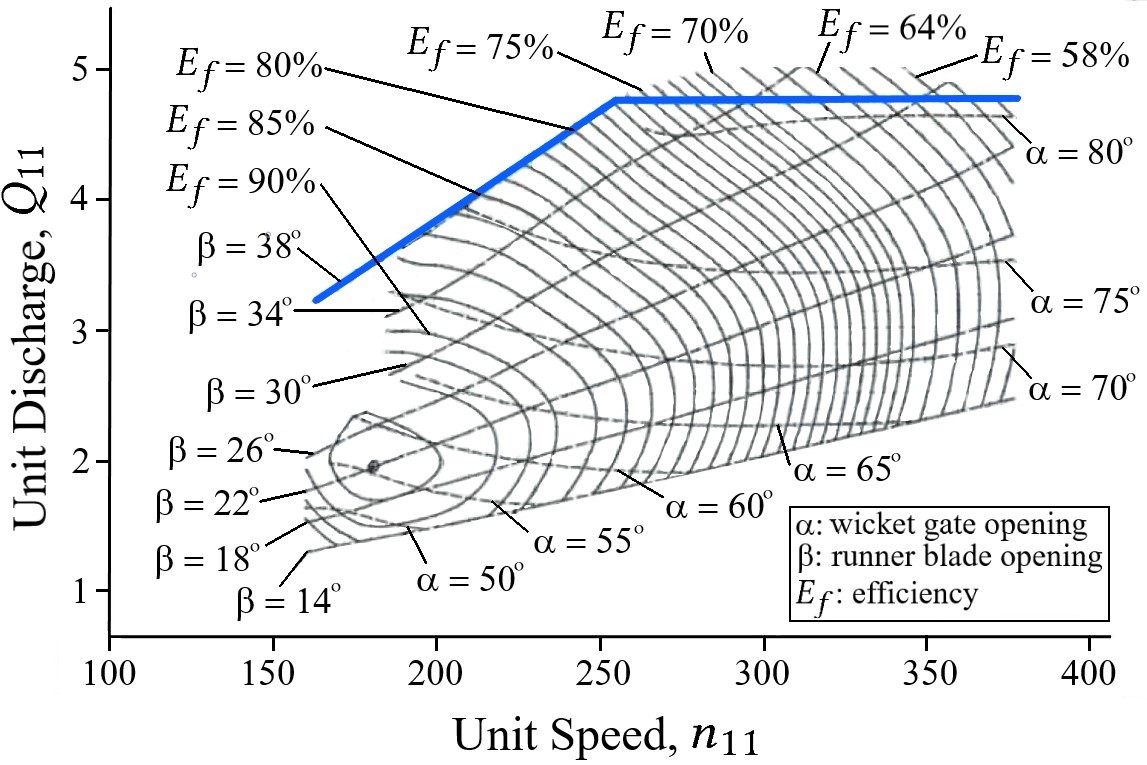}
	\caption{Edited Andritz Chart for a double regulated turbine (varying $\alpha$ and $\beta$ angles), adapted from \cite{aggidis2012tidal}. The blue line represents the parameterized maximum power output curve.}
	\label{ANDRITZ} 
\end{figure}

From 0D to 2D models, studies show that flow rate and power from turbines can be approximated with the parametrization of experimental results \cite{falconer2009severn,aggidis2013operational,aggidis2012tidal}. For this work, the equations describing flow and power for low head bulb turbines were based on experimental results from Andritz Hydro
\cite{aggidis2012tidal}. The edited Andritz chart shown in Fig. \ref{ANDRITZ} demonstrates how turbine unit speed $n_{11}$ and specific unit discharge $Q_{11}$ (obtained experimentally) are related. The graph also shows wicket gate and running blade openings ($\alpha$ and $\beta$, in degrees), and iso-efficiency curves $E_{f}$.

By specifying the parameters of the turbine: diameter $D$, number of generating poles $G_p$ and grid frequency $f$, the turbine rotation $S_p$ ($rpm$) is obtained from $S_p = 120f/G_p$. Furthermore, unit speed $n_{11}$, turbine flow rate $Q_t$ and power output $P$ are calculated as:
\begin{equation}
n_{11} = S_p D / \sqrt{|H_e|},
\label{n11}
\end{equation}
\begin{equation}
Q_t = Q_{11}D^2\sqrt{|H_e|},
\label{Qt}
\end{equation}
\begin{equation}
P = |\rho g Q_t H_e E_{f} C_E|.
\label{P}
\end{equation}

$H_e$ is the head difference between ocean and lagoon, $\rho$ the seawater density ($1024~kg/m^3$), $g$ the gravity acceleration ($9.81~m/s^2$) and $C_E$ is the product of other efficiencies shown in Table \ref{Efficiencies}.

\begin{table}
\centering
  \caption{Other efficiency considerations for TRS \cite{aggidis2013operational}.}\label{Efficiencies} % or \label{TotalLosses}
  \begin{tabular}{ccl}
    \toprule
    TRS Efficiencies & (\%) \\
    \midrule
    Generator  & 97 \\
    Transformer  & 99.5 \\
    Water friction  & 95 \\
    Gear box/drive train  & 97.2 \\
    Turbine availability  & 95 \\
    Turbine orientation (Flood Gen. only) \cite{angeloudis2018optimising} & 90 \\
  \bottomrule
\end{tabular}
\end{table}

When $H_e$ is available, $n_{11}$ is estimated directly from Eq.~(\ref{n11}). For calculating $Q_t$ and $P$, $Q_{11}$ and $E_f$ are obtained experimentally by adjusting the opening of the wicket gates ($\alpha$), the pitch angle of the runner blades ($\beta$) and crossing the values with the obtained $n_{11}$ (see Fig. \ref{ANDRITZ}). In order to choose appropriate values for $\alpha$ and $\beta$, a parameterized curve of maximum power output was drawn over Fig. \ref{ANDRITZ} (blue line) by following the path where the product between $E_f$, $Q_t$ and $H_e$ is maximised. If we assume $\alpha$ and $\beta$ are automatically adjusted to always be in the maximum power output curve, then $Q_{11}$ and $E_f$ become functions of $n_{11}$, as shown in Eq.~(\ref{Q112}) and (\ref{etaz}):
\begin{equation}
\label{Q112}
\begin{aligned}
Q_{11} &= (0.0166)n_{11} + 0.4861;\ (\text{when} \ n_{11} \leqslant 255)\\
Q_{11} &= 4.75;\ (\text{when} \ n_{11} > 255),
\end{aligned}
\end{equation}
and
\begin{equation}
E_f = (-0.0019)n_{11} + 1.2461.
\label{etaz}
\end{equation}

For simulating sluice gates, the barrage model utilises the orifice equation, so that the flow rate $Q_o$ is a function of $H_e$ \cite{prandle1984simple,Baker}:
\begin{equation}
Q_o = C_dA_S\sqrt{2g|H_e|},
\label{Orif}
\end{equation}
where $C_d$ is the discharge coefficient for sluices (equal to one in this study, following \cite{angeloudis2018optimising}), and $A_S$ the sluice area.

When generating energy, turbines use Eq.~(\ref{Qt}) for estimating flow rate through the barrage. On the other hand, when operating in ``idling'' mode, turbines use equation Eq.~(\ref{Orif}), with $C_d = 1.36$ \cite{angeloudis2018optimising}. 

When starting or stopping either turbines or sluices, the literature has used sinusoidal ramp functions $r(t)$ for simulating the smooth transition of flow output as $r(t) = sin[(\pi/2) (t - t_m)/t_r]$ \cite{angeloudis2018optimising, angeloudis2017comparison}, where $t \in [t_m, t_m + t_r]$, $t_r = \text{transition time}$ (around 15 $min$ to 20 $min$ \cite{angeloudis2018optimising, angeloudis2016numerical}), and $t_m$ is the time when the current operation was triggered. Since this is a heuristic method, a simpler transition function (named ``momentum ramp'') is proposed in this work:
\begin{equation}
Q_{t+1} = Q_c - \zeta \Delta Q.
\label{TransitionTulio}
\end{equation}
$Q_{t+1}$ is the estimated flow rate at the next time-step, $Q_c$ is the total flow rate calculated from turbines and sluice equations (Eq.~(\ref{Qt}, \ref{Orif})), $\zeta$ a dimensionless hyperparameter that controls the intensity of flow rate update per time-step, $Q_{t}$ is the flow rate at time-step ``$t$'' and $\Delta Q = Q_c - Q_t$.

The ``momentum ramp'' is applied every time-step during simulation. This not only simplifies the code, but facilitates training, since sluice opening is treated as a continuous control problem (Section \ref{AgentEnv}). In this work we set $\zeta = .4$, which guarantees a precision of $10^6$ for a 15 $min$ time interval with $\Delta t = 60s$.

% Fig.~\ref{NovelRamp} compares results from the momentum ramp function with a sinusoidal 15 min ramp function. 

% \begin{figure}
% 	\centering
% 	\includegraphics[width=\linewidth]{samples/RampFunctions.PNG}
% 	\caption{Comparison between classic sinusoidal and novel ``momentum'' ramp function ($\zeta = .4$ and $\Delta t = 60s$) when Starting hydraulic structures.}
% 	\label{NovelRamp} 
% \end{figure}

% The `momentum'' ramp function also has an advantage when considering the operation of independent groups of turbines, since $Q_c$ is simply a function of the number of operating units at a certain time-step.

% A more thorough analysis of the barrage boundary implementation and verification (considering power generation) can be found in \cite{andrea2019implementation}.

\section{Reinforcement Learning  Overview}

\begin{figure}[h]
	\centering
	\includegraphics[width=.7\linewidth]{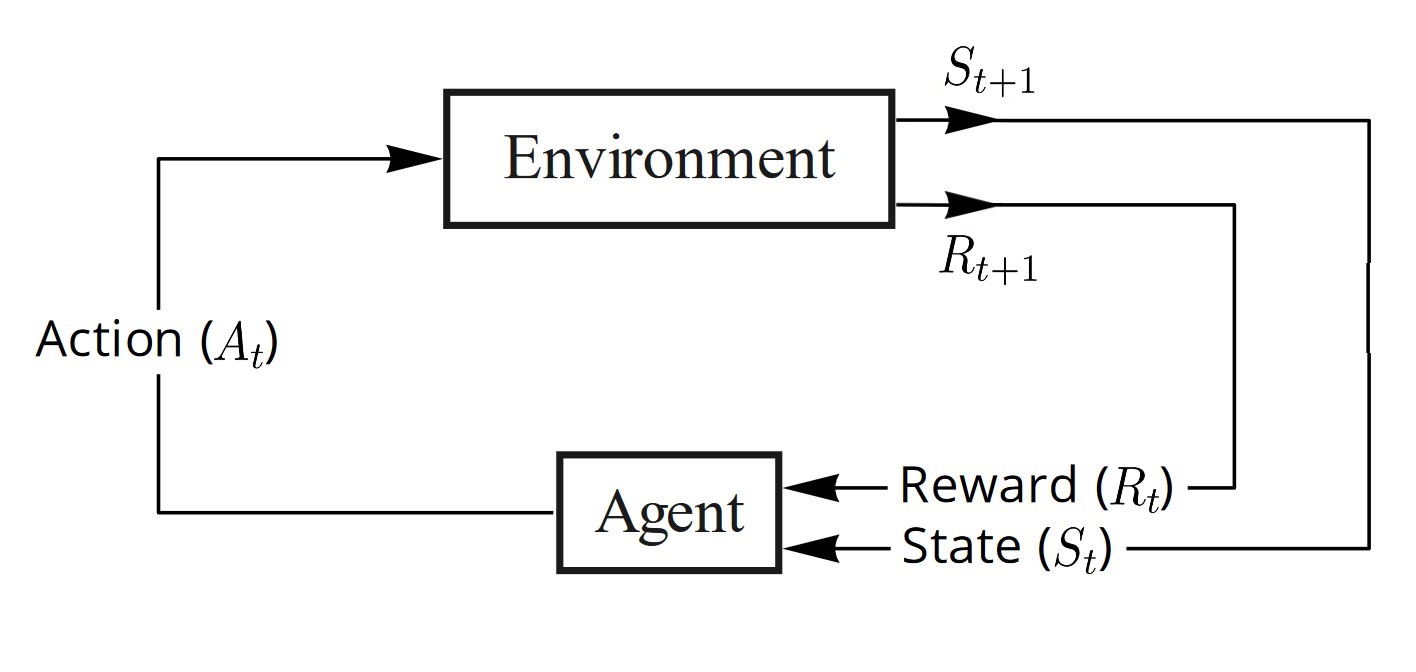}
	\caption{Agent-environment interaction in an MDP, illustrating the state-action, next reward triples sequence. Adapted from \cite{sutton2018reinforcement}.}
	\label{RLFramework} 
\end{figure}

As shown in the work of Sutton and Barto \cite{sutton2018reinforcement}, a reinforcement learning (RL) problem can be mathematically formalised as a Markov Decision Process (MDP). In an MDP, an agent interacts with an environment through actions ($A_t$), and these actions lead to new environmental states ($S_{t+1}$) and possible rewards ($R_{t+1}$) for the agent. The quantities $A_t$, $S_t$ and $R_t$ are random variables, with well-defined probability distributions. A general agent-environment MDP framework can be visualised in Fig. \ref{RLFramework}.

By sampling multiple time-steps $t = 0, 1, 2, 3 ...$, observations ($O_i$) of the agent-environment interaction are organised as a sequence of state-action, next reward triples:
\begin{equation}
O_i = < s_i, a_i, r_{i+1} >,
\label{Observation}
\end{equation}
where, $s_i, a_i, r_{i+1}$ are instances of the random variables ($S_t$, $A_t$ and $R_t$). The sequence of state-action pairs defines a trajectory $\tau$:
\begin{equation}
\tau = s_0, a_0, s_1, a_1, s_2, a_2 ...
\label{Trajectory}
\end{equation}

Also, in an MDP, we can say that the probabilities of $R_{t+1}$ and $S_{t+1}$ are completely conditioned on the preceding state and action ($S_t$ and $A_t$), that is:
\begin{equation}
p(s_{t+1}, r_{t+1}|s_t, a_t).
\label{Dynamics}
\end{equation}

The probability distribution of Eq.~(\ref{Dynamics}) defines the dynamics of the MDP. It can also be manipulated to yield the state-action-transition probability distribution (which is just the sum of probabilities over all possible future rewards):
\begin{equation}
p(s_{t+1}|s_t, a_t) = \sum_{r \in R_{t+1}} p(s_{t+1}, r_{t+1}|s_t, a_t).
\label{StateActionTransition}
\end{equation}

For estimating Eq.~(\ref{StateActionTransition}) for a given state, we also need to condition an action. In non-deterministic scenarios, the selection of possible actions by the agent is a stochastic process, defined by a conditional probability distribution (known as policy) of the form:
\begin{equation}
\pi(a_t|s_t).
\label{Policy}
\end{equation}

Using Eq.~(\ref{StateActionTransition},\ref{Policy}), the probability distribution of starting in a state $s_t$ and ending in $s_{t+1}$, given a policy, can be estimated as:
\begin{equation}
p_{\pi}(s_{t+1}|s_t) = \sum_{a \in A_t} \pi(a_t|s_t) p(s_{t+1}|s_t, a_t).
\label{StateTransitionFull}
\end{equation}

% For a single path $i$:

% \begin{equation}
% p_{\pi}(s_{t+1}^{(i)}|s_t^{(i)}) = \pi(a_t^{(i)}|s_t^{(i)}) p(s_{t+1}^{(i)}|s_t^{(i)}, a_t^{(i)}).
% \label{StateTransition}
% \end{equation}

With a defined policy, we can sum the observed rewards for each state-action pair (as shown in Eq.~(\ref{Observation})) and calculate a total return $G_t$ at time-step $t$:
\begin{equation}
G_t = R_{t+1} + \gamma R_{t+2} + \gamma^2 R_{t+3} ... = \sum^\infty_{k=0} \gamma^k R_{t+k+1},
\label{Return}
\end{equation}
where $\gamma$ is a discount factor between 0 and 1.

The objective of reinforcement learning problems is to find an optimal policy $\pi^*$, that maximises the expected return of rewards $E[G_t]$ conditioned on any initial state, i.e.
\begin{equation}
\pi^* = \argmax_{\pi} E_{\pi} [G_t | S_t = s_t], \forall s_t.
\label{GoalPolicy}
\end{equation}

% Finally, the expected reward, following the optimal policy, can be conditioned on a state or state-action pair, yielding state value and action value functions, respectively:

% \begin{equation}
% V_{\pi^*} (s_t) = E_{\pi^*} [G_t | s_t]
% \label{VFunction}
% \end{equation}

% and

% \begin{equation}
% Q_{\pi^*} (s_t, a_t)= E_{\pi^*} [G_t | s_t, a_t].
% \label{QFunction}
% \end{equation}

\subsection{Proximal Policy Optimisation (PPO)}
\label{PPOSection}

Once the reinforcement learning problem is formalised as an MDP, several algorithms can be used for finding an optimal control policy $\pi^*$. In this work, the process of finding $\pi^*$ has been achieved through Proximal Policy Optimisation (PPO) \cite{schulman2017proximal}, built in the Unity ML-Agents package. PPO was shown to outperform several other ``on-policy'' gradient methods \cite{schulman2017proximal} and is one of the preferred methods for control optimisation when the cost of acquiring new data is low \cite{lecturePieter2}. Furthermore, a revised and up-to date PPO algorithm is already implemented in Unity ML-Agents. Nevertheless, for sake of completeness, we present in this section the mathematical derivation of the PPO method. The designing of the TRS operation as an MDP is shown in Section \ref{AgentEnv}.

Differently from approaches that try to infer the policy through state-value or action-value functions (e.g. Deep Q-Network) \cite{mnih2013playing}, PPO uses an ``on-policy'' approach that maximises the expected sum of rewards by improving its current policy -- smoothly shifting the probability density function estimate of the policy towards $\pi^*$. The PPO algorithm is an updated form of Policy Gradients. As TRPO (Trust Region Policy optimisation)  \cite{schulman2015trust}, it tries to increase sample efficiency (re-using data from previous policies), while constraining gradient steps to a trust region. It is also actor-critic, since it utilises an estimate of the state-value function for its baseline \cite{lecturePieter3}. An overview of Policy Gradients and PPO is presented below.

\subsubsection{Policy Gradients}

\begin{figure}
	\centering
	\includegraphics[width=.7\linewidth]{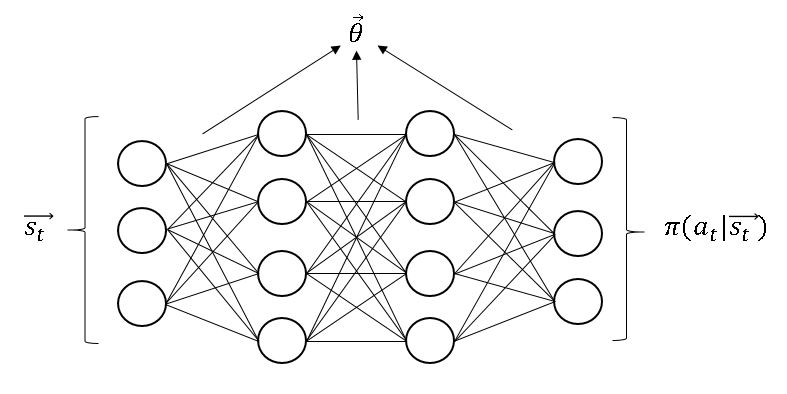}
	\caption{Input-output representation of policy (actor) neural network.}
	\label{PolicyNetwork} 
\end{figure}

Policy gradient methods rely on the fact that a stochastic policy can be parameterized by an ``actor'' neural network with weights $\Vec{\theta}$ (simplified as $\theta$ going forward). As represented in Fig. \ref{PolicyNetwork}, this neural network receives a vector state representation of $s_t$. For the case of discrete actions, the neural network outputs the probabilities of each possible action in that state using a softmax layer. For continuous actions, each node in the last layer outputs the moments of a multivariate Gaussian distribution of the form \cite{lecturePoupart,bohn2019deep}:
\begin{equation}
\pi_{\theta} (a_t|s_t) = N(a_t|\mu(s_t;\theta), \Sigma(s_t;\theta)),
\label{policyContinuous}
\end{equation}
where $\mu$ and $\Sigma$ are the parameterized mean and co-variance matrices, respectively. While training, actions are randomly sampled from the distribution to favour exploration. During testing, $\mu$ is taken as the optimum action for each input state $s_t$.

Considering a trajectory $\tau$, the expected return of following a parameterized policy $\pi_{\theta}$ is $U(\theta) =  E [G_{(\tau)} ; \pi_{\theta}]$, where
\begin{equation}
G_{(\tau)} = \sum_{t = 0}^H R(s_t, a_t),
\label{ReturnTarget}
\end{equation}
and $R(s_t, a_t)$ is the reward from taking action $a_t$ from state $s_t$. We also note that $G_{(\tau)}$ represents the undiscounted return following a sampled trajectory $\tau$ for a time horizon $H$. With these considerations, finding an optimal policy can be viewed as tuning $\theta$ to maximise $U(\theta)$, i.e. to perform gradient ascent of $\theta$:
\begin{equation}
\theta \leftarrow \theta + \alpha \nabla_{\theta} U(\theta).
\label{GradAscent}
\end{equation}

A sample based estimate for $\nabla_{\theta} U(\theta)$ assumes the form:
\begin{equation}
\nabla_{\theta} U(\theta) \approx \frac{1}{m} \sum_{i = 1}^m \sum_{t = 0}^{H} \nabla_{\theta} log (\pi_{\theta}(a_t^{(i)}|s_t^{(i)}))  \hat{A}_t,
\label{GradExpectedR4}
\end{equation}
where $m$ is the number of sampled trajectories from the ``actor'' neural network. For vanilla policy gradient methods \cite{ lecturePieter}, following a trajectory $\tau^{(i)}$, we get
\begin{equation}
\hat{A}_t = G_{(\tau)}^{future} - V_{\pi}(s_t^{(i)}), \quad \text{and}
\label{GradExpectedR2}
\end{equation}
\begin{equation}
G_{(\tau)}^{future} = \sum_{k = t}^{H} \gamma^{(k-t)} R(s_k^{(i)}, a_k^{(i)}).
\label{GradExpectedR3}
\end{equation}

$V_{\pi}(s_t^{(i)})$, parameterized by a ``critic'' neural network, is the estimate for the value function of being in state $s_t$ and following policy $\pi$ thereafter; $G_{(\tau)}^{future}$ is the discounted future return of following the chosen trajectory, from time $k=t$; and $\hat{A}_t$ is the advantage estimate of taking this trajectory in respect to the current estimate of $V_{\pi}(s_t^{(i)})$. A complete derivation of $\nabla_{\theta} U(\theta)$ can be seen in \cite{lecturePieter}.

\subsubsection{Clipped Surrogate Loss derivation for PPO} \label{SurrLoss}

In order to increase sampling efficiency \cite{lectureSchulman}, importance sampling can be used to rewrite the gradient term in Eq.(~\ref{GradExpectedR4}) as:
\begin{equation}
\hat{g}^{IS} = \hat{E_t} \left[\nabla_{\theta} \frac{\pi_{\theta}(a | s)}{\pi_{\theta old} (a | s)} \hat{A}_t \right]
\label{GradExpectedR6}.
\end{equation}

Eq.(~\ref{GradExpectedR6}) importance sampling form allows for re-utilising samples from an older policy to perform gradient ascent steps, when refining a new policy. It is obtained when differentiating the Surrogate Loss:
\begin{equation}
L_{\theta old}^{IS}(\theta) = \hat{E}_{t} [r_t (\theta) \hat{A}_t ],
\label{SurrogateLoss}
\end{equation}
where $r_t (\theta) = \pi_{\theta}(a | s) / \pi_{\theta old} (a | s)$ is a probability ratio.

While in TRPO \cite{schulman2015trust} the maximization of the surrogate loss from Eq.(~\ref{SurrogateLoss}) is subjected to a Kullback–Leibler divergence constraint, in PPO \cite{schulman2017proximal} the surrogate loss is constrained through a clipping procedure, yielding the clipped surrogate loss objective:
\begin{equation}
L^{CLIP} (\theta) = \hat{E}_t [min(r_t(\theta) \hat{A}_t, clip(r_t (\theta), 1 - \epsilon, 1 + \epsilon) \hat{A}_t )],
\label{Clipped}
\end{equation}
where $\epsilon$ is a hyperparameter that limits large policy updates.

To further reduce variance when estimating the advantage, \cite{schulman2017proximal, lecturePieter3} utilise a truncated version of generalized advantage estimation \cite{schulman2015high}, where $\hat{A}_t$ is estimated as 
\begin{equation}
\hat{A}_t = \delta_t + (\gamma \lambda)\delta_{t+1} + ... + (\gamma \lambda)^{H-t+1} \delta_{H-1},
\label{AtGAE}
\end{equation}
where $t$ is a time index within the sampled trajectory time horizon $[0, H]$, $\lambda$ is a hyperparameter that performs the exponential weighted average of k-step estimators of the returns \cite{schulman2015high}, and $\delta_t = r_t + \gamma V(s_{t+1}) - V(s_t)$.

When utilising shared parameters for the ``actor'' and ``critic'' neural networks (as is the case with this work), the loss function needs to be augmented with a value function error term \cite{schulman2017proximal}. To ensure exploration, an entropy term ``$S$'' is also added. Finally, the loss function to be maximised at each iteration becomes: 
\begin{equation}
L_t^{CVS} (\theta) = \hat{E}_t [L_t^{CLIP} (\theta) - c_1 L_t^{VF} (\theta) + \beta S[\pi_{\theta}] (s_t)].
\label{Clipped2}
\end{equation}

For this study, Unity ML-Agents package fixes $c_1 = 0.5$ \cite{Pierre2020}, $\beta$ is a hyperparameter controlling the entropy bonus magnitude $S$, and $L_t^{VF}$ is a clipped, squared-error loss between the estimate of the state-value function $V_{\pi}(s_t;\theta)$ and the actual return value obtained when following a trajectory $\tau$ \cite{schulman2017proximal}. The implementation of $L_t^{VF}$ in Unity ML-Agents is seen in \cite{Pierre2020}.

Additionally, parallel training can also be implemented as a way of substituting experience replay by running the policy on multiple instances of the environment. By guaranteeing that each environment starts in a random initial state during training, this parallelism helps decorrelate the sampled data, stabilising learning \cite{mnih2016asynchronous}.

\section{Agent-Environment Setup}

\subsection{Unity ML-Agents}

The Unity3D graphics engine is a popular game developing environment that has been used to create games and simulations in 2D and 3D since its debut in 2005. It has received
widespread adoption in other areas as well, such as architecture, engineering and construction \cite{juliani2018unity}.

Unity ML-Agents is an open-source project that allows for designing environments where a smart agent can learn through interactions \cite{juliani2018unity, Pierre2020}. It has been chosen in this project due to ease of implementation, built-in PPO algorithm and visual framework for visualising real-time control of TRS.

\subsection{Agent-Environment MDP Modelling and Training}
\label{AgentEnv}
To formalise TRS operation as a RL problem (and subsequently solve the RL problem through PPO) we need to design an MDP in Unity ML-Agents with environment, agent, actions, states and reward components.

\begin{figure}
  \centering
  \begin{minipage}[b]{.6\textwidth}
    \includegraphics[width=1\linewidth]{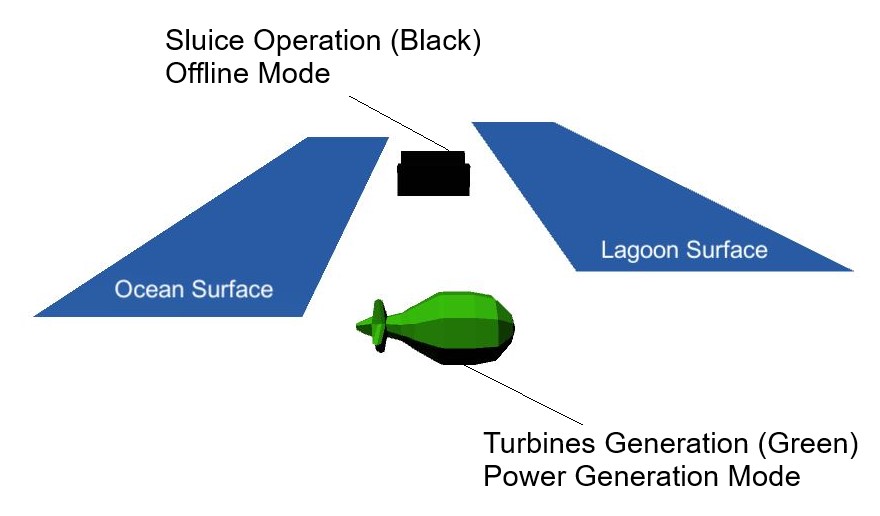}
  \end{minipage}
  \caption{Unity ML-Agents MDP environment for a 0D model of the Swansea Bay Tidal Lagoon during ebb generation.} \label{EbbGenUnity}
\end{figure}

By creating simple representative 3D models for turbines, sluices, ocean and lagoon, a training environment simulating TRS for our MDP is created in Unity3D and then imported to a Unity ML-Agents project. In this environment, the equations for simulating flow operation through the lagoon, when operating sluices and turbines, are extracted from the 0D model representation, detailed in Section \ref{Equations}. In order to choose appropriate parameters for operating our environment, we follow literature representations suggested by \cite{angeloudis2018optimising, angeloudis2017comparison, xue2019optimising}, for the Swansea Bay Tidal Lagoon project. The chosen parameters are shown in Table\,\ref{Swansea_P}. A variable lagoon surface area, digitized from \cite{xue2019optimising}, is also utilised. For ease of visualisation, the 3D representations of sluice and turbine change colours depending on the operational mode chosen by the agent. For the turbine, green represents power generation mode, orange -- idling mode and black -- offline mode (zero flow rate). Similarly, sluices change colour between orange and black for sluicing and offline modes, respectively. Fig. \ref{EbbGenUnity} shows a capture of the Unity ML-Agents MDP environment representation for the Swansea Bay tidal Lagoon during ebb generation, with the representative models for sluice and turbines in offline and power generation modes, respectively. Ocean and Lagoon surface level motion are also represented.

\begin{table}[h]
\centering
      \caption{Swansea Lagoon design.}\label{Swansea_P}
  \begin{tabular}{ccl}
  \toprule
	$n^o$ of Turbines & $16$  \\
	$N^o$ of $Gp$ & $95$  \\
	Grid frequency $(Hz)$ & $50$  \\
	Turbine Diameter ($m$) & $7.35$  \\
	Sluice Area ($m^2$) & $800$  \\
  \bottomrule
\end{tabular}
\end{table}

Additionally, in this proposed MDP, the actor-critic agent is defined as an operator responsible for controlling turbine and sluice operational modes through actions (policy network node outputs $n_{o}$), according to a vector of input states $s_t$. $n_{o}$ outputs can be discrete or continuous. In this work, continuous outputs are chosen, reducing the number of nodes in the last layer, and consequently, the complexity of the neural network. There are $3$ node outputs that determine turbine and sluice operation every 15 $min$ of the environment simulation. The 15 $min$ window (MDP time-step) was selected for this work since the time usually associated with the opening/closing of hydraulic structures lies in the $[\text{15 }min - \text{20 }min]$ range \cite{angeloudis2018optimising, angeloudis2016numerical}.

Each node in the last layer outputs a value between $0$ and $1$, and the resulting actions are computed in a hierarchical fashion. The first node determines the number of turbines set to power generation mode ($0$ or $16$), depending if the node output is below or above a threshold ($0.5$), i.e. if the node outputs a value below the threshold, no turbines will be generating energy, otherwise, all $16$ turbines are set to power generating mode. Therefore, if no turbines are set to power generation, $16$ are available for other operational modes (idling or offline).

The second node selects the number of idling turbines just as the first node, if the number of turbines available is $16$. Otherwise the number of idling turbines is $0$, independent of this node output. If no turbine is selected for power generation or idling modes, all turbines are set offline. Therefore, the first two nodes control turbines through discrete actions.

The third and final node outputs the $\%$ opening area of the sluice gates. Since any value between $[0,1]$ can be chosen by the neural network, the momentum ramp function (Section \ref{Equations}) is applied to the outputted flow rate every time-step, ensuring smooth flow rate transitions, independently of the opening sluice area set by the agent.

Beyond reducing the number of node outputs, this configuration also allows for having the sluice operation independent of turbine operation. All possible operational modes for turbines and sluices as a function of node output ($n_o$) are shown in Tables \ref{TurbConfig} and \ref{SluiceConfig}, respectively.

\begin{table}[h]
\centering
  \caption{Possible turbine operational modes.}\label{TurbConfig}
  \begin{tabular}{cccl}
    \toprule
	Node 1 & Node 2 & Discrete Turbine Control \\
    \midrule
	$n_{o1} < 0.5$ & $n_{o2} < 0.5$ & Offline Mode\\
	$n_{o1} < 0.5$ & $n_{o2} \geq 0.5$ & Idling Mode\\
	$n_{o1} \geq 0.5$ & $n_{o2} < 0.5$ & Power Generation Mode\\
	$n_{o1} \geq 0.5$ & $n_{o2} \geq 0.5$ & Power Generation Mode\\
  \bottomrule
\end{tabular}
\end{table}

\begin{table}[h]
\centering
  \caption{Possible sluice operational modes.}\label{SluiceConfig}
  \begin{tabular}{ccl}
    \toprule
	Node 3 & Continuous Sluice Control \\
    \midrule
	$0 < n_{o3} \leq 1$ & Sluicing Mode (Available sluice area = $n_{o3}A_S$)\\
	$n_{o3} = 0$ & Offline Mode (Available sluice area = 0)\\
  \bottomrule
\end{tabular}
\end{table}

The actions selected by the agent are a function of the input states $s_t$. In this work, these states are the water levels of ocean and lagoon, plus current operational mode of turbines and sluices, for current and previous MDP time-steps (Table \ref{States}). Finally, the reward received by the agent equals the accumulated energy generated by the turbines, for every 15 $min$.

\begin{table}[h]
\centering
  \caption{Input states for PPO neural network.}\label{States}
  \begin{tabular}{ccl}
    \toprule
	States (at times $t$ and $t-1$)  & Units  \\
    \midrule
	Ocean water level & meters (float) \\
	Lagoon water level & meters (float) \\
	Number of online turbines & 0, or 16 (integer) \\
	Number of idling turbines &  0, or 16 (integer) \\
	Sluice gate opening area &  0 to 1 (float) \\
  \bottomrule
\end{tabular}
\end{table}

\begin{figure}
	\centering
	\includegraphics[width=\columnwidth]{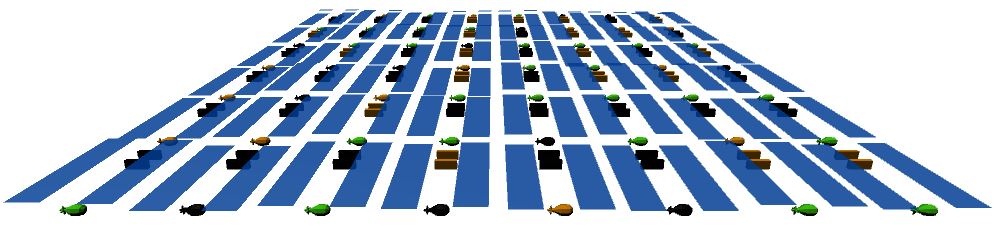}
	\caption{64 instances of the environment during parallel training. For the turbines, green represents power generation mode. For turbines and sluices orange represents idling/sluicing mode and black -- offline mode.}
	\label{ParallelTraining} 
\end{figure}

For stabilising and speeding up training, parallel training is performed with 64 copies of the environment (Fig. \ref{ParallelTraining}), while episodes are set to $1$ month of duration. During training, each environment instance requires a representative ocean input at the location where the Swansea Lagoon is planned to be constructed ($51^{\circ}35'58.9"N$ $3^{\circ}53'42.4"W$). Ideally, ocean measurements could be used as training data. However, due to the lack of sufficient measured data (Section \ref{baselines}), it is not possible to train the agent until reasonable performance is reached. Instead, an artificial tide signal to simulate the ocean is created by summing the major sinusoidal tide constituents (due to gravitational pull of the Moon and Sun). Although we are not accounting for other less predictable local wave motions (e.g. wind waves), the artificial ocean input representation is sufficient for enabling the agent to converge to an optimal policy. A major advantage of this approach is the fact that we can generate any amount of input data required for training the agent.

The tide constituent's amplitudes of the simulated ocean utilised in this work (Table \ref{TideConstituents}), were obtained from a numerical simulation, at the location of Swansea Bay Tidal Lagoon, by \cite{angeloudis2018optimising}. The periods for each constituent were obtained by \cite{wolanski2015estuarine}. The final equation for simulating the ocean can be seen in Eq.~(\ref{OceanModel}).
\begin{equation}
\label{OceanModel}
\begin{aligned}
o(t) = A_{M2} sin(\omega_{M2}t + \phi_{M2}) + A_{S2} sin(\omega_{S2}t + \phi_{S2}) + & \\ 
A_{N2} sin(\omega_{N2}t + \phi_{N2}) + A_{K1} sin(\omega_{K1}t + \phi_{K1}),  & 
\end{aligned}
\end{equation}

where $\omega_{M2}$, $\omega_{S2}$, $\omega_{N2}$ and $\omega_{K1}$ are angular frequencies (rad/s) of each tidal component, and $\phi_{M2}$, $\phi_{S2}$, $\phi_{N2}$ and $\phi_{K1}$ are random phase lags in the range $[0,2\pi]$, generated for each environment instance during parallel training when starting an episode, which allow for learning more generalised scenarios. 

\begin{table}[h]
\centering
  \caption{Simulated tide constituents at Swansea Bay.}\label{TideConstituents}
  \begin{tabular}{cccl}
    \toprule
    Ocean tide constituent  & Amplitude (m) & Period (hr)  \\
    \midrule
	$M2$  & $A_{M2} = 3.20$ & $T_{M2} = 12.42$ \\
	$S2$  & $A_{S2} = 1.14$ & $T_{S2} = 12$ \\
	$N2$  & $A_{N2} = 0.61$ & $T_{N2} = 12.66$ \\
	$K1$  & $A_{K1} = 0.08$ & $T_{K1} = 23.93$ \\
  \bottomrule
\end{tabular}
\end{table}

\begin{figure}[h]
	\centering
	\includegraphics[width=.9\columnwidth]{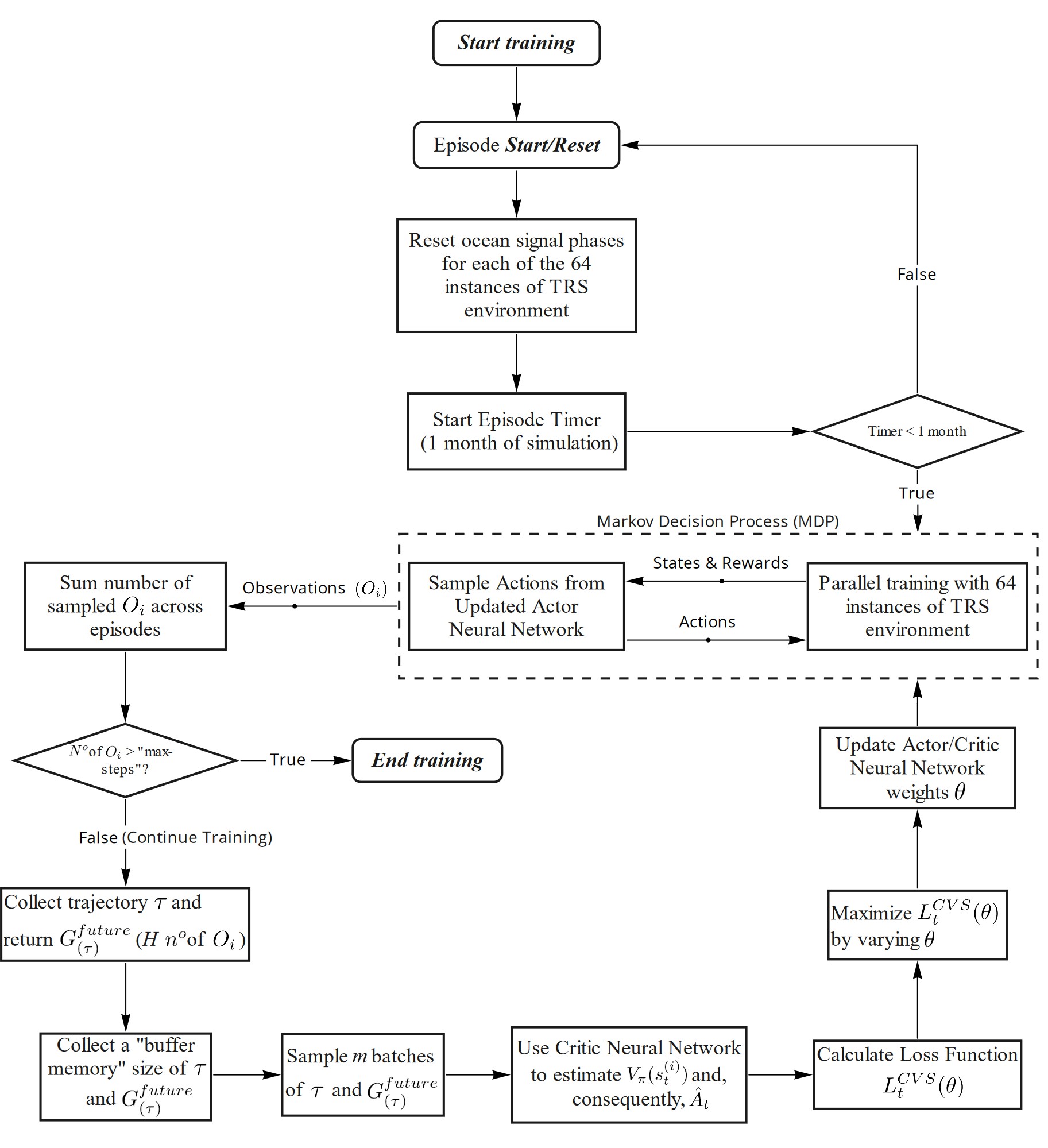}
	\caption{Flow-chart with detailing of the PPO algorithm in Unity ML-Agents, during training. For a given trajectory $\tau^{(i)}$, $\hat{A}_t$ is the advantage estimate and $V_{\pi}(s_t^{(i)})$ the value function.}
	\label{TrainBlock} 
\end{figure}

The designed MDP is solved through the PPO algorithm (Section \ref{PPOSection}). A flow-chart, illustrating the agent's training stage through the PPO algorithm in Unity ML-Agents, is presented in Fig. \ref{TrainBlock}. Training occurs until a ``max-steps'' number of observations are sampled. For reproducibility, Table S2 in the Supplementary Material showcases the PPO hyperparameters utilised during training.

After training, the policy (actor) neural network receives input states $s_t$ and outputs optimum $n_o$ values, following a policy that maximises energy generation. During testing, this means that the agent receives real ocean measurements as input, performing real-time flexible control of turbines and sluices.

\section{Experiments} \label{Results}

In this section we compare our DRL trained agent performance against state-of-the-art optimisation routines. Codes for reproducibility can be made available under request to the corresponding author.

\subsection{Test Data and Baselines optimisation}
\label{baselines}

For comparing our DRL agent performance against conventional optimisation routines, we model six baselines devised from the recent literature \cite{angeloudis2018optimising, xue2019optimising} and compare the energy generated in a month for each method. All baselines in this work consider the operation of the Swansea Bay tidal lagoon either through classic or variant ``two-way scheme'' methods, as detailed in Section \ref{TidalPower}.

Regarding test data for baselines and trained agent, we utilise all tide gauge ocean measurements available from the British Oceanographic Data Centre (BODC) at Mumbles Station \cite{bodc}, located at the edge of Swansea Bay. The obtained measurements of ocean elevation are recorded every 15 $min$ in a table, for the years of $1993$ and the range $[1997 - 2019]$. Before utilising the data, a preprocessing step is performed so that data flagged as ``improbable'', ``null value'' and ``interpolated'' by BODC are not considered. After this step we retain $26$ months of usable, non-overlapping, test data. The preprocessing step ensures a conservative comparison between baselines and our trained agent, since it considers scenarios where tidal predictions had a good match with measured data.

Tidal predictions for the same 26 months are also provided by BODC in the same data-set. For each month, baseline optimisation routines utilise tidal predictions for capturing operational head values $H_{start}$, $H_{min}$ and $HS_{start}$ (when considered) that optimise power generation. These operational head values are then applied to the measured ocean test data, so that comparisons between baselines and trained agent can be made. Baselines, in increasing order of optimisation complexity, are described next:

\begin{itemize}
    \item CH (Constant Heads): Best, constant $H_{start}$ and $H_{min}$ are picked for extracting energy during a whole month \cite{ahmadian2017optimisation}.
    
    \item CHV (Constant Heads, with variant operation): Best, constant heads $H_{start}$, $H_{min}$ and $HS_{start}$ are picked for extracting energy during a whole month.
    
    \item EHT (Every Half-Tide): optimised pairs of $H_{start}$ and $H_{min}$ are picked for every consecutive half-tide. Proposed by \cite{xue2019optimising}.
    
    \item EHTV (Every Half-Tide, with variant operation): optimised $H_{start}$, $H_{min}$ and $HS_{start}$ are picked for every consecutive half-tide.
    
    \item EHN (Every Half-Tide and Next): optimised $H_{start}$ and $H_{min}$ are picked for every half-tide, considering the best $H_{start}$ and $H_{min}$ for the next half-tide as well. Proposed by \cite{xue2019optimising}.
    
    \item EHNV  (Every Half-Tide and Next, with variant operation): optimised $H_{start}$, $H_{min}$ and $HS_{start}$ are picked for every half-tide, considering the best $H_{start}$, $H_{min}$ and $HS_{start}$ for the next half-tide as well.
\end{itemize}
All variant optimisation methods are augmented through the addition of independent sluice head operation $HS_{start}$. This modification we are introducing is inspired by the work of \cite{Baker, angeloudis2018optimising}. CH and CHV perform non-flexible operation, while EHT, EHTV, EHN and EHNV perform state-of-art flexible operation. A summary detailing each baseline operational heads and method is shown in Table \ref{tableBaselines}. 

\begin{table} [h]
  \scriptsize
  \centering
  \caption{Simplified reference table for baselines.}\label{tableBaselines}
\begin{tabular}{@{}lllllll@{}}
\toprule
        & \multicolumn{2}{l}{Constant Head} & \multicolumn{2}{l}{Every Half-Tide} & \multicolumn{2}{l}{Every Half-Tide and Next} \\ \midrule
        & CH          & CHV            & EHT         & EHTV          & \quad EHN         & \quad \quad EHNV     \\
$H_{start}$  & \ \bcheck   & \ \ \bcheck    & \ \ \bcheck & \quad \bcheck & \quad \ \ \bcheck & \quad \quad \quad  \bcheck \\
$H_{min}$    & \ \bcheck   & \ \ \bcheck    & \ \ \bcheck & \quad \bcheck & \quad \ \ \bcheck & \quad \quad \quad  \bcheck  \\
$HS_{start}$ &             & \ \ \bcheck    &             & \quad \bcheck &                   & \quad \quad \quad  \bcheck  \\
$\text{non-flexible operation}$ & \ \bcheck   & \ \ \bcheck    &             &             &             & \quad \quad \quad  \\
$\text{flexible operation}$ &             &             & \ \ \bcheck & \quad \bcheck & \quad \ \ \bcheck & \quad \quad \quad  \bcheck  \\ \bottomrule
\end{tabular}
\end{table}

% \begin{table} [h]
%   \scriptsize
%   \centering
%   \caption{Simplified reference table for baselines.}\label{tableBaselines}
% \begin{tabular}{@{}lllllll@{}}
% \toprule
%         & \multicolumn{2}{l}{Constant Head} & \multicolumn{2}{l}{Every Half-Tide} & \multicolumn{2}{l}{Every Half-Tide and Next} \\ \midrule
%         & CH          & CHV            & EHT         & EHTV          & \quad EHN         & \quad \quad EHNV     \\
% $Classical Operation$  & \ \bcheck   & \ \ \bcheck    & \ \ \bcheck & \quad \bcheck & \quad \ \ \bcheck & \quad \quad \quad  \bcheck \\
% $Variant Operation$    & \ \bcheck   & \ \ \bcheck    & \ \ \bcheck & \quad \bcheck & \quad \ \ \bcheck & \quad \quad \quad  \bcheck  \\
% $Non-Flexible Operation$ &             & \ \ \bcheck    &             & \quad \bcheck &                   & \quad \quad \quad  \bcheck  \\ \bottomrule
% \end{tabular}
% \end{table}

All baselines, except EHNV, are optimised with a grid search optimisation algorithm, which iteratively increases its search resolution until convergence. Initial search resolution starts with $1$ meter, with optimisation heads $H_{start}$, $H_{min}$ and $HS_{start}$ (when considered) within ranges $[1m - 6m]$, $[1m - 3m]$ and $[1m - 5m]$, respectively. After the first run, search resolution is halved and the algorithm performs a brute-force search around the best previous configuration attained. The latter procedure is repeated until final search resolution is lower than $1cm$.

EHNV requires a different optimisation approach due to its high computational time when utilising the previous grid search method. For this case we utilise the stochastic global optimisation algorithm basin-hopping \cite{wales1997global} from Scipy package \cite{2020SciPy}, with COBYLA as a local minimizer \cite{powell1994advances}. Basin-hopping was chosen for its efficiency when solving smooth function problems with several local minima separated by large barriers \cite{basinho}. The local minimizer COBYLA is a nonlinear derivative–free constrained optimisation that uses a linear approximation approach. Even though basin-hopping is not guaranteed to converge to a global optimum, EHNV is shown to be, on average, the best baseline method for energy generation.

\subsection{Agent Performance Evaluation} \label{Results2}

\begin{figure}[h]
	\centering
	\includegraphics[width=.7\linewidth]{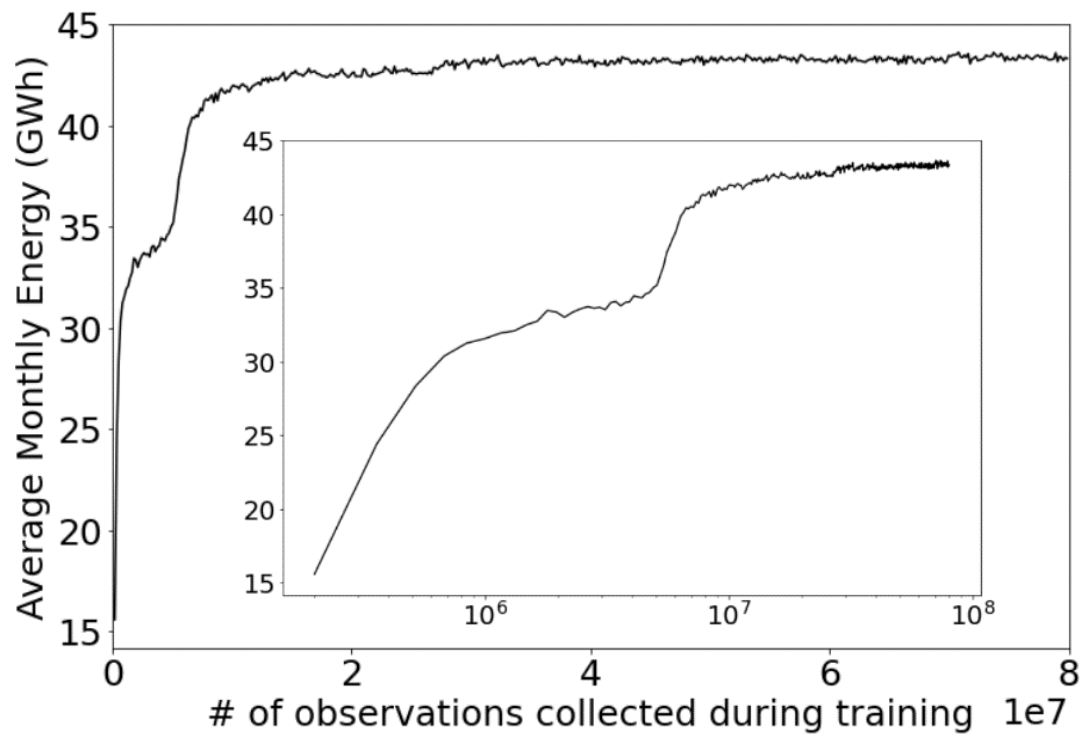}
	\caption{Monthly cumulative energy (reward) in GWh, averaged for all 64 environments during parallel training. The log-representation insert highlights the two-step plateau.}
	\label{FinalR} 
\end{figure}

Following hyperparameter tuning, we trained the agent for $8E7$ steps, until convergence. The cumulative reward (energy) per month (episode) during parallel training, averaged for the 64 instances of the lagoon environment, is shown in Fig. \ref{FinalR}. The log-representation insert highlights the two-step plateau that is observed when converging to an optimal strategy. After starting in a total random strategy, the cumulative reward received by the agent increases until reaching an intermediate plateau at around $2E6$ steps, where the agent learns the strategy of operating mostly the turbines, while keeping sluices practically offline during ebb generation. Then, after about $5E6$ steps, the cumulative reward starts increasing again. The second plateau stabilises around $4E7$ steps, with a cumulative reward approximately 25\% higher than the first plateau -- a gain allowed by (i) a flexible operational strategy learnt by the agent, that adjust TRS operation according to tidal range (ii) the smart usage of the sluicing mode, as discussed below in test results. Videos showcasing the strategic operation developed for both plateaus are available in the Supplementary Material. 

\begin{figure}[h]
	\centering
	\includegraphics[width=.5\linewidth]{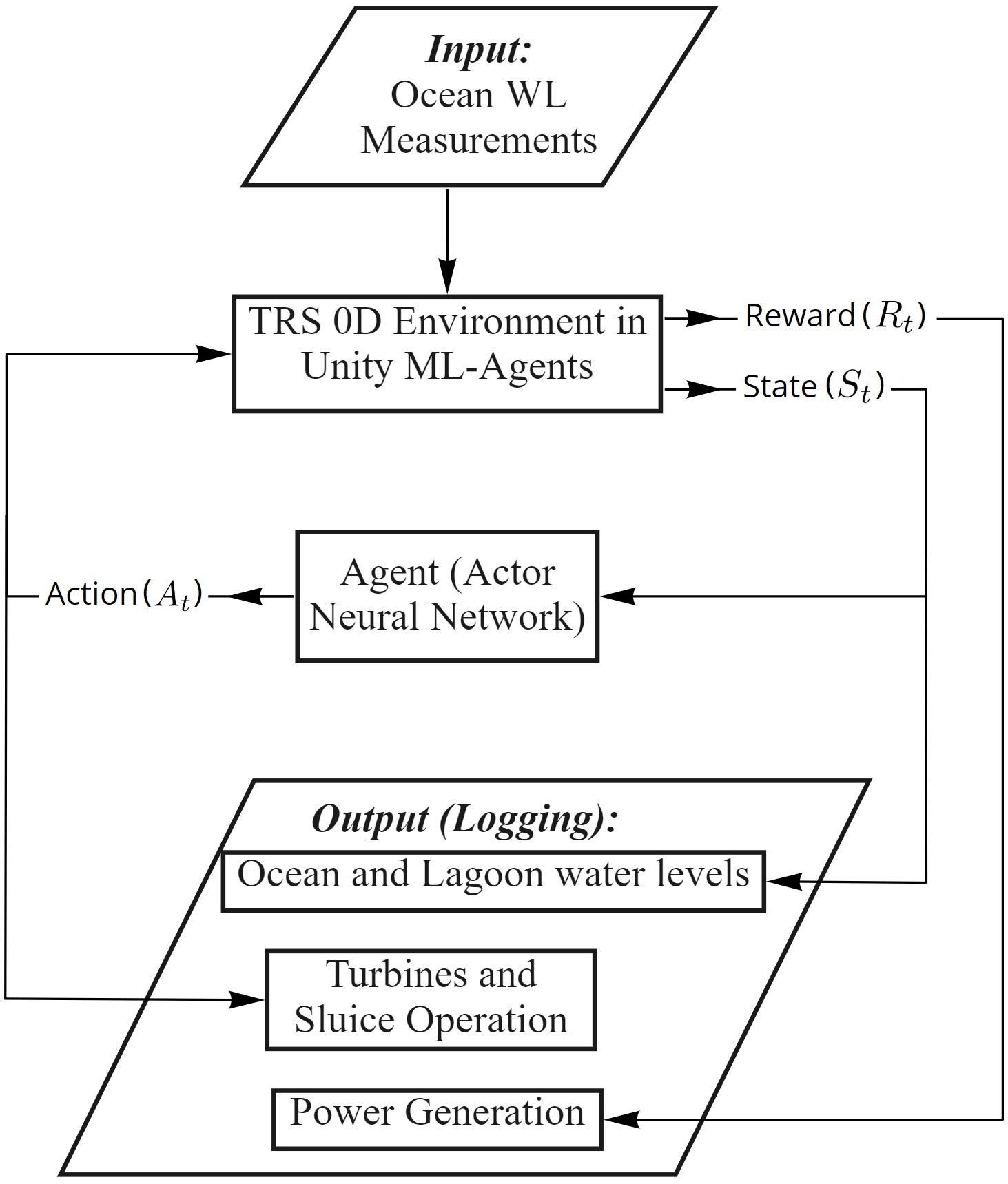}
	\caption{Block diagram with detailing of the trained agent test stage.}
	\label{TestBlock} 
\end{figure}

\begin{figure}[h]
	\centering
	\includegraphics[width=.6\linewidth]{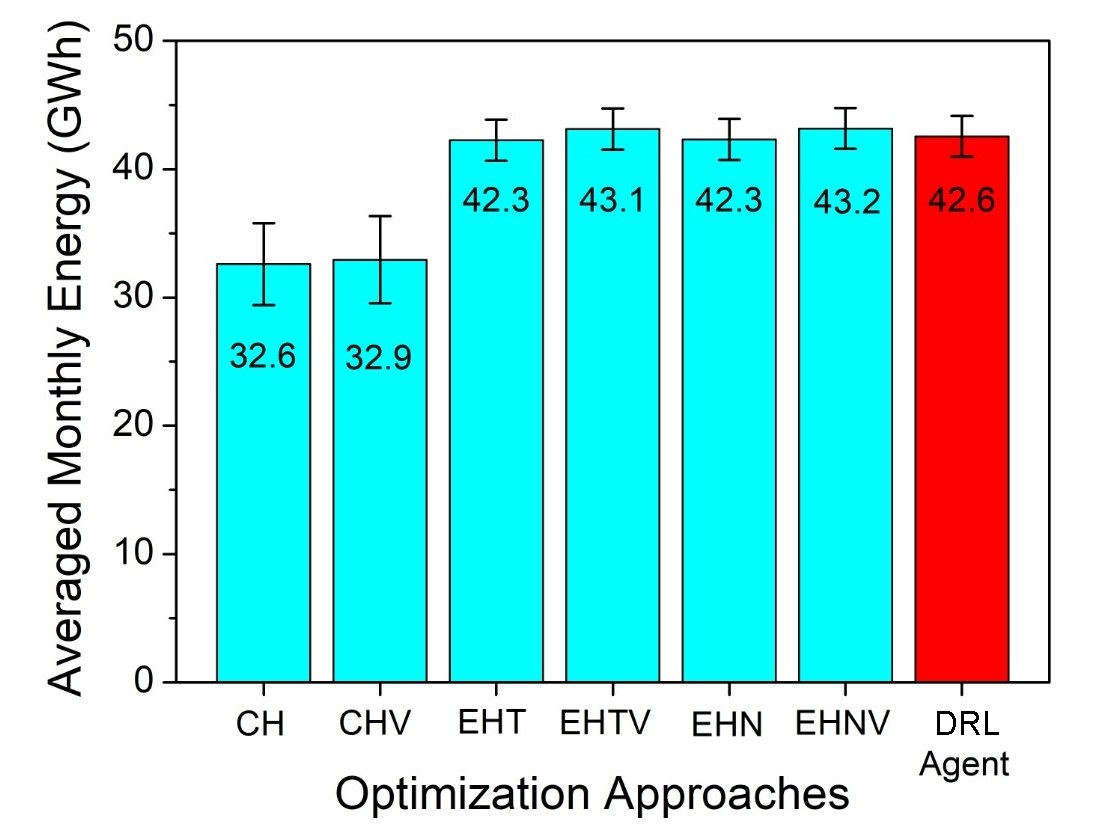}
	\caption{Averaged monthly energy comparison between baselines and trained agent utilising test data. Sample standard deviations for the various months are also shown as error bars.}
	\label{Barplot} 
\end{figure}

For test data, we utilise $26$ months of real ocean measurements from BODC. These months are presented and numbered in Table S3, while Table S4 (Supplementary Material) compares the amount of energy obtained in the numbered months between our trained agent (performing real-time flexible control) and the baselines. A block diagram representing the trained agent test stage for this work, while using ocean water level (WL) measurements, is shown in Fig. \ref{TestBlock}. The averaged monthly energy attained for all methods is shown in Fig. \ref{Barplot}.

For the baselines, CH and CHV present the worst performance, since constant operational heads cannot account for the varying ocean amplitudes in a month (about $\approx \text{2 }m$ to $\approx \text{4.5 }m$ in our test set). Furthermore, baselines with variant operation outputted more energy in average than their classical counterparts. Finally, ``half-tide and next'' approaches showed very small improvements ($<0.2\%$) when compared to ``half-tide'' methods, while requiring much greater ($\gtrapprox 20 \times$) computational time (Tables S5 and S6, Supplementary Material).

\begin{figure*}
  \centering
  \begin{subfigure}[t]{.49\linewidth}
    \centering\includegraphics[width=.95\linewidth]{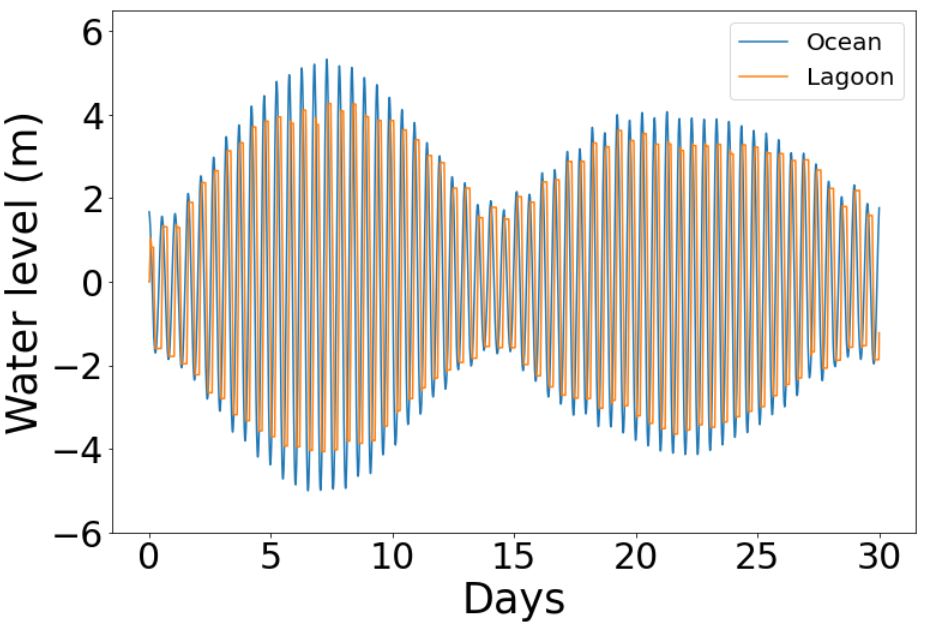}
    \caption{Ocean and lagoon water levels.} \label{AgentOL}
  \end{subfigure}
  \begin{subfigure}[t]{.49\linewidth}
    \centering\includegraphics[width=.95\linewidth]{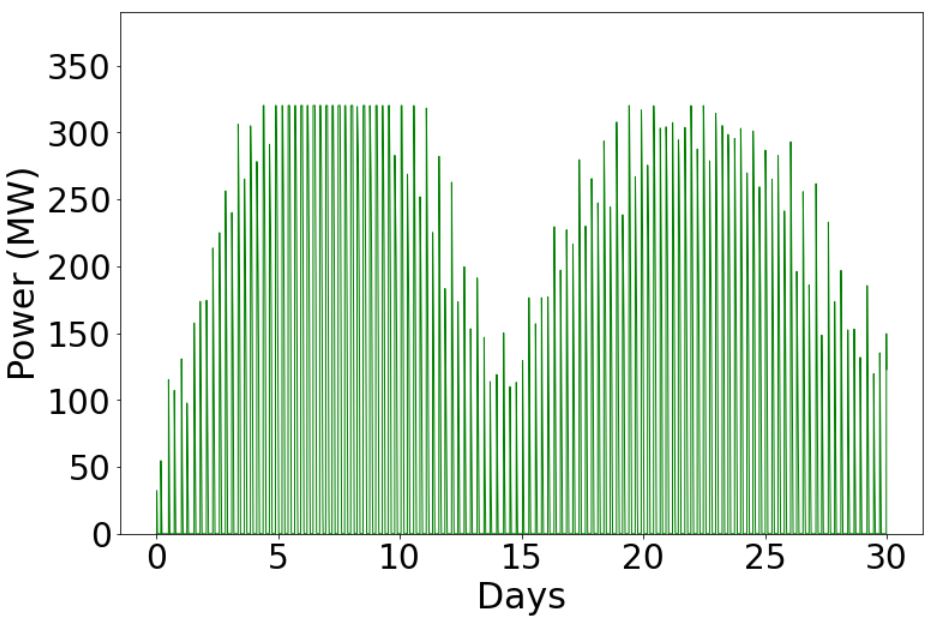}
    \caption{Power generation.}
    \label{AgentPower}
  \end{subfigure}
  \caption{Lagoon water levels and power generation results for the trained agent performing flexible control in a month, with measured ocean data only.} \label{FlexControl}
\end{figure*}

\begin{figure*}
  \centering
  \begin{subfigure}[t]{.48\linewidth}
    \centering\includegraphics[width=.95\linewidth]{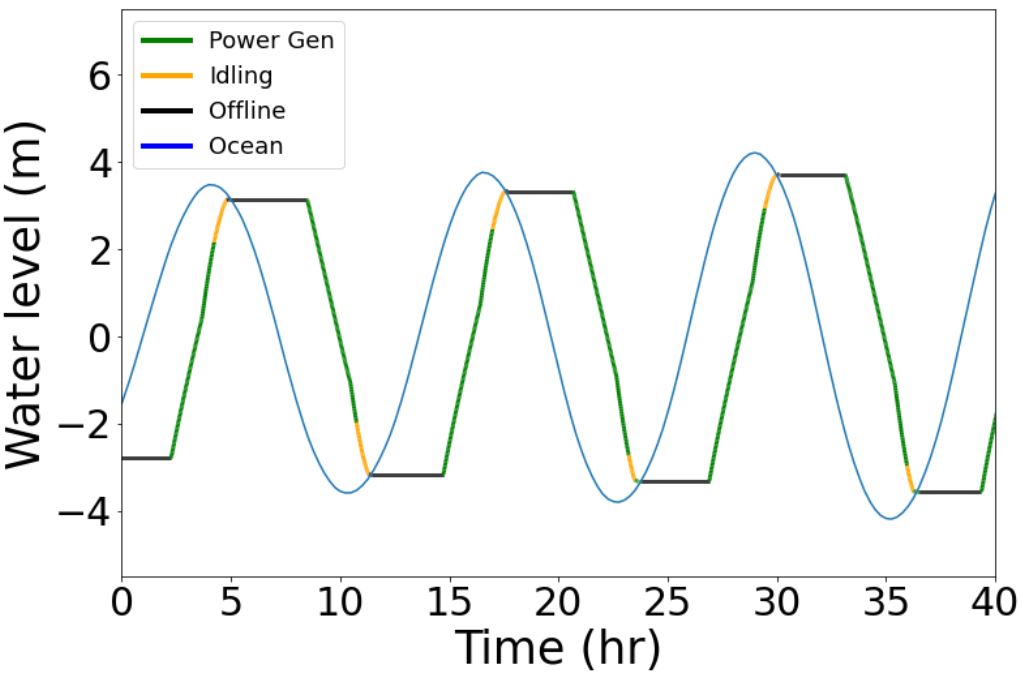}
    \caption{Ocean (in blue) and lagoon water levels.} \label{AgentTurbineWl}
  \end{subfigure}
  \begin{subfigure}[t]{.51\linewidth}
    \centering\includegraphics[width=.95\linewidth]{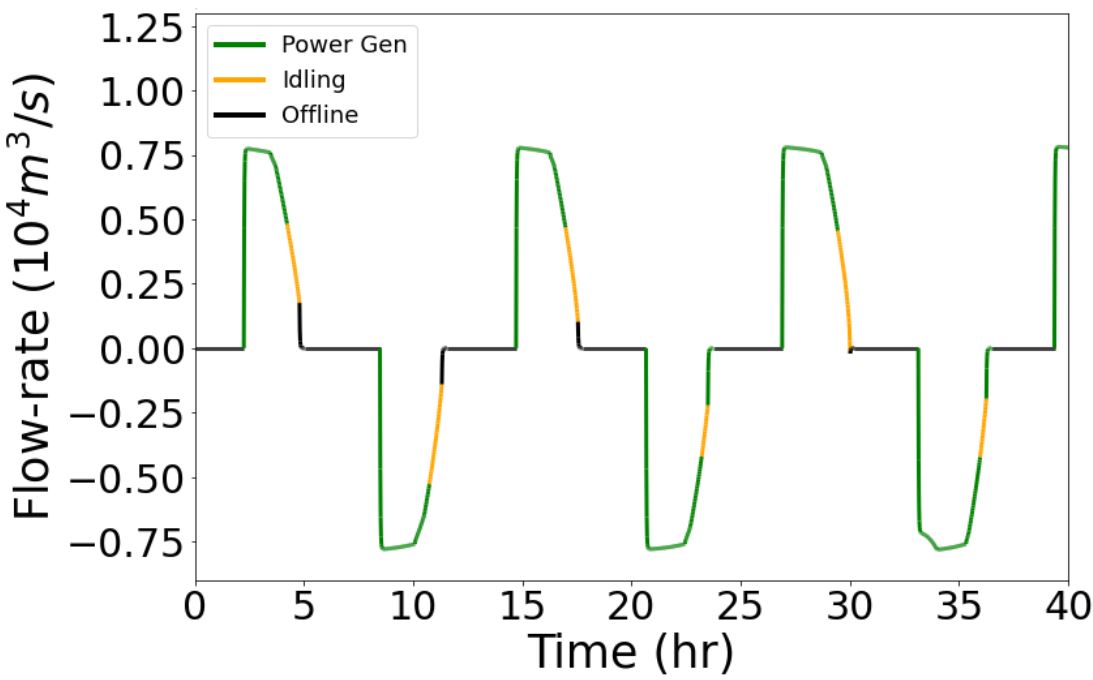}
    \caption{Flow rate from the 16 turbine units.}
    \label{AgentTurbineQt}
  \end{subfigure}
  \begin{subfigure}[t]{.49\linewidth}
    \centering\includegraphics[width=.95\linewidth]{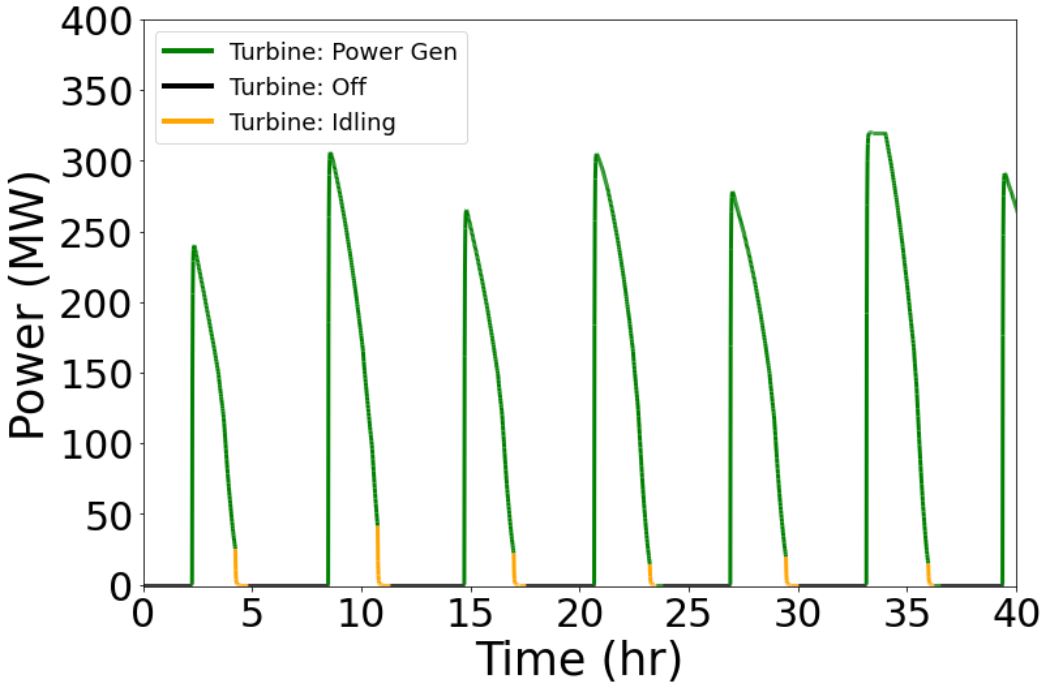}
    \caption{Combined power output from turbines.} \label{AgentTurbinePow}
  \end{subfigure}
  \caption{Lagoon water levels, turbine flow rates and power output are shown and coloured following turbine operational mode chosen by the trained agent. Green represents power generation mode, orange -- idling mode and black -- offline mode.} \label{turbineAgent}
 \end{figure*}

\begin{figure*}
  \centering
  \begin{subfigure}[t]{.485\linewidth}
    \centering\includegraphics[width=.95\linewidth]{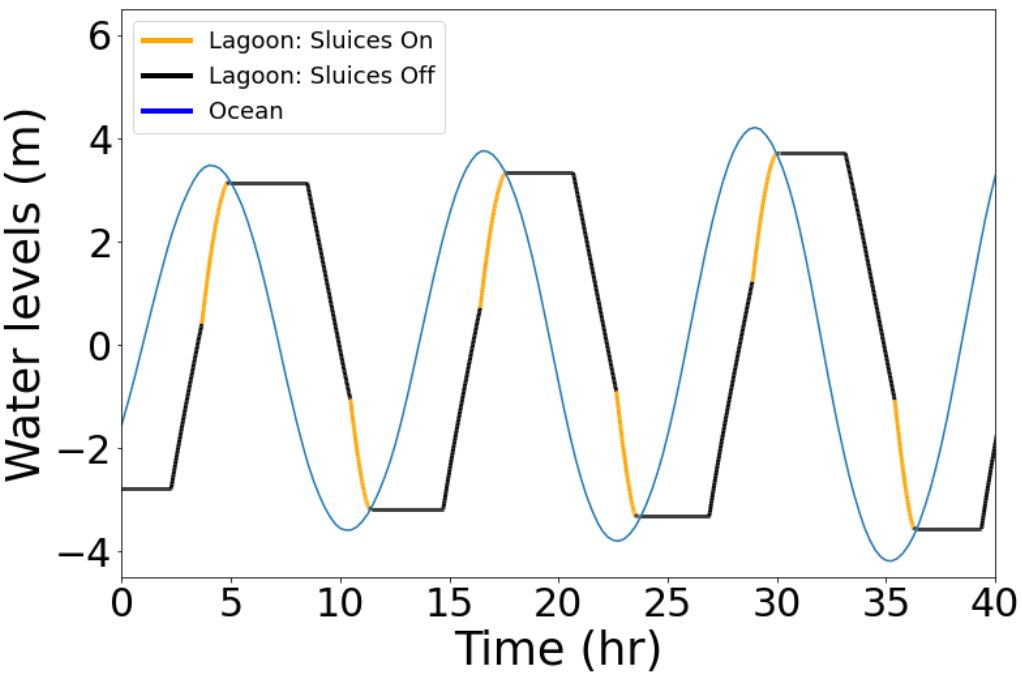}
    \caption{Ocean (in blue) and lagoon water levels.} \label{AgentSluiceWl}
  \end{subfigure}
  \begin{subfigure}[t]{.495\linewidth}
    \centering\includegraphics[width=.95\linewidth]{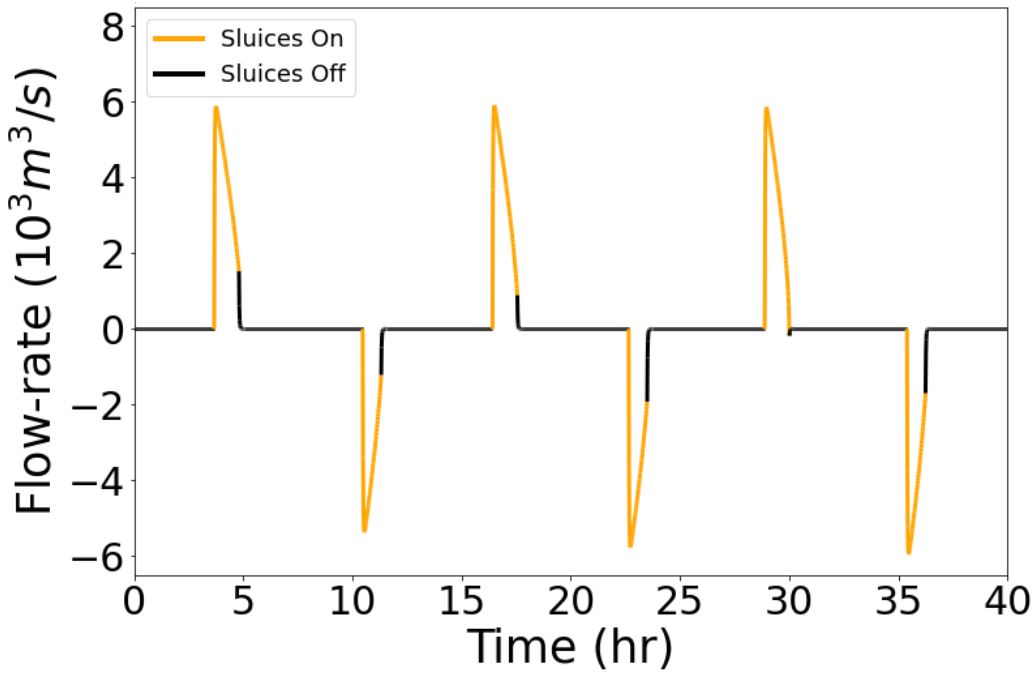}
    \caption{Flow rate from sluice gates.}
    \label{AgentSluiceQs}
  \end{subfigure}
  \caption{Lagoon water levels and sluice flow rates are shown and coloured following sluice operational mode chosen by the trained agent. Orange represents idling (i.e. sluicing) mode and black -- offline mode.} \label{sluiceAgent}
\end{figure*}

For the trained agent, Fig. \ref{FlexControl} show operational test results of power generation and lagoon water levels for one month of measured ocean data (starting with initial lagoon water level at mean sea level). We note that the agent quickly converges to an optimal energy generation strategy for sequential tidal cycles, independent of tidal range input -- a characteristic of state-of-art flexible operation \cite{xue2019optimising}. Furthermore, Figs. \ref{turbineAgent} and \ref{sluiceAgent} showcase detailed results of real-time control on test data. Apart from ocean water levels, results are coloured according to actions taken by the agent for turbines and sluices, respectively, as defined in Section \ref{AgentEnv}. More specifically, Figs. \ref{AgentTurbineWl} and \ref{AgentSluiceWl} show lagoon water level variations, while Fig. \ref{AgentTurbinePow}, \ref{AgentTurbineQt} and \ref{AgentSluiceQs} show power generation, turbine and sluice flow rates. From the sequence of actions taken, we see that the agent arrives at a policy with independent operation of sluices, i.e. the variant operation of TRS, which was shown to be a better strategy than the classical operation in our baseline comparison. A summary of our method accomplishments in comparison with state-of-art baselines is shown in Table \ref{ComparisonTable}.

\begin{table}
\setlength{\tabcolsep}{3pt}
\scriptsize
\centering
\caption{Comparison of state-of-the-art baselines with our proposed DRL Agent.}
\begin{tabular}{@{}llllllll@{}}
\toprule
                         & CH & CHV$^a$ & EHT     & EHTV$^a$ & EHN     & EHNV$^a$ & {\bf DRL Agent}     \\
                         & \cite{ahmadian2017optimisation} & & \cite{xue2019optimising} & &  \cite{xue2019optimising} & & {\bf (our work)}  \\ \cmidrule(l){2-8} 
real-time flexible control        &    &             &             &               &             &               & \quad \quad  \bcheck             \\
prediction-free approach &    &             &             &               &             &               & \quad \quad  \bcheck              \\
variant lagoon operation &    & \ \ \bcheck &             & \quad \bcheck &             & \quad \bcheck & \quad \quad  \bcheck             \\
state-of-art performance$^b$ &    &             & \ \ \bcheck & \quad \bcheck & \ \ \bcheck & \quad \bcheck & \quad \quad  \bcheck             \\ \midrule
\multicolumn{8}{l}{$^a$ optimisation routines with variant operation of tidal lagoons. Augmentation inspired by \cite{angeloudis2018optimising, Baker}} \\
\multicolumn{8}{l}{$^{b}$ Equivalent outputs, in average, within the error bars (Fig. \ref{Barplot}).}                                                                             \label{ComparisonTable}        

\end{tabular}
\end{table}

Our agent managed very competitive energy outputs, staying on average within $1.4\%$ of the best baseline (EHNV). Indeed, for all months tested, our agent performed optimally, outputting better results than the state-of-art EHN method for 22 out of 26 months (Table S4, Supplementary Material), within a $1.1\%$ margin in worst scenarios. We note that this novel result was obtained by training the agent once with a simple artificial ocean input, in contrast with the baselines that require future tidal predictions and being re-run for every new tide.

\section{Conclusions}

In this work, we have shown that Proximal Policy Optimisation (a DRL method) can be used for real-time flexible control of Tidal Range Structures after training with artificially generated tide signals. Our DRL agent implementation was then compared against state-of-art optimisation approaches devised from the literature, yielding competitive results for all test data utilised. Our results were obtained in a conservative setting, i.e., when available tidal measurements had good agreement with tidal predictions (a requirement for state-of-art approaches).

We have chosen the Swansea Bay Tidal Lagoon for our analysis, given its status as a pathfinder project for larger tidal lagoon projects. We show that our novel approach obtains optimal energy generation from measured tidal data only, through an optimised control policy of turbines and sluices, which are operated independently by the trained DRL agent. Our method shows promising advancements over state-of-art optimisation approaches since it (i) performs real-time flexible control with equivalent energy generation, (ii) does not require future tidal predictions, and (iii) needs to be trained a single time only.

Owing to its characteristic features, the method introduced here shall be of wide applicability to achieve optimal energy generation of TRS in cases where future tidal predictions are unreliable or not available. Possible extensions of this work include more sophisticated cases, such as through the introduction of pumping capabilities and/or demand-oriented operation (where the goal is maximising revenue). For such cases, the increase in both the degrees of freedom for the agent's actions and stochasticity from the input signal could further emphasise the advantage of a DRL real-time operation of TRS over current state-of-art methods.

% For future work, we will investigate the generalisation capabilities of our method for other tidal lagoons. Furthermore, allowing turbines to operate as pumps  can further increase power output of usual optimisation methods up to $\approx 25\%$ \cite{xue2019optimising}. Efforts for introducing pumping in our approach are in progress and will be reported elsewhere.

\section*{Acknowledgments}
We would like to thank the Brazilian agencies Coordenação de Aperfeiçoamento de Pessoal de Ensino Superior (CAPES) and Conselho Nacional de Desenvolvimento Científico e Tecnológico (CNPq) for providing funding for this research.

\bibliography{mybibfile}

\end{document}